\documentclass[10pt,twocolumn,letterpaper]{article}

\usepackage{cvpr}
\usepackage{times}
\usepackage{epsfig}
\usepackage{graphicx}
\usepackage{amsmath}
\usepackage{amssymb}
\usepackage{bm}
\usepackage{multirow}
\usepackage{nicefrac}       
\usepackage{algorithm, algpseudocode}
\usepackage{fixltx2e}
\usepackage{array}

\DeclareMathOperator*{\argmax}{arg\,max} 
\DeclareMathOperator*{\argmin}{arg\,min} 

\usepackage{color}

\let\oldReturn\Return
\renewcommand{\Return}{\State\oldReturn}

\usepackage[pagebackref=true,breaklinks=true,letterpaper=true,colorlinks,bookmarks=false]{hyperref}

\cvprfinalcopy 

\def\cvprPaperID{4098} 

\begin{document}

\title{Boosting Adversarial Attacks with Momentum}

\author{Yinpeng Dong$^{1}$, Fangzhou Liao$^{1}$, Tianyu Pang$^{1}$, Hang Su$^{1}$, Jun Zhu$^{1}$\thanks{Corresponding author.}, Xiaolin Hu$^{1}$, Jianguo Li$^{2}$\\
$^{1}$ Department of Computer Science and Technology, Tsinghua Lab of Brain and Intelligence\\
$^{1}$ Beijing National Research Center for Information Science and Technology, BNRist Lab\\
$^{1}$ Tsinghua University, 100084 China\\
$^{2}$ Intel Labs China\\
\small{\{dyp17, liaofz13, pty17\}@mails.tsinghua.edu.cn, \{suhangss, dcszj, xlhu\}@mail.tsinghua.edu.cn}, \small{jianguo.li@intel.com}
}

\maketitle

\begin{abstract}

Deep neural networks are vulnerable to adversarial examples, which poses security concerns on these algorithms due to the potentially severe consequences. Adversarial attacks serve as an important surrogate to evaluate the robustness of deep learning models before they are deployed. However, most of existing adversarial attacks can only fool a black-box model with a low success rate. To address this issue, we propose a broad class of momentum-based iterative  algorithms to boost adversarial attacks. By integrating the momentum term into the iterative process for attacks, our methods can stabilize update directions and escape from poor local maxima during the iterations, resulting in more transferable adversarial examples. To further improve the success rates for black-box attacks, we apply momentum iterative algorithms to an ensemble of models, and show that the adversarially trained models with a strong defense ability are also vulnerable to our black-box attacks. We hope that the proposed methods will serve as a benchmark for evaluating the robustness of various deep models and defense methods. With this method, we won the first places in NIPS 2017 Non-targeted Adversarial Attack and Targeted Adversarial Attack competitions. 
\end{abstract}

\section{Introduction}

\begin{figure}[!t]
  \centering
    \includegraphics[width=1.0\linewidth]{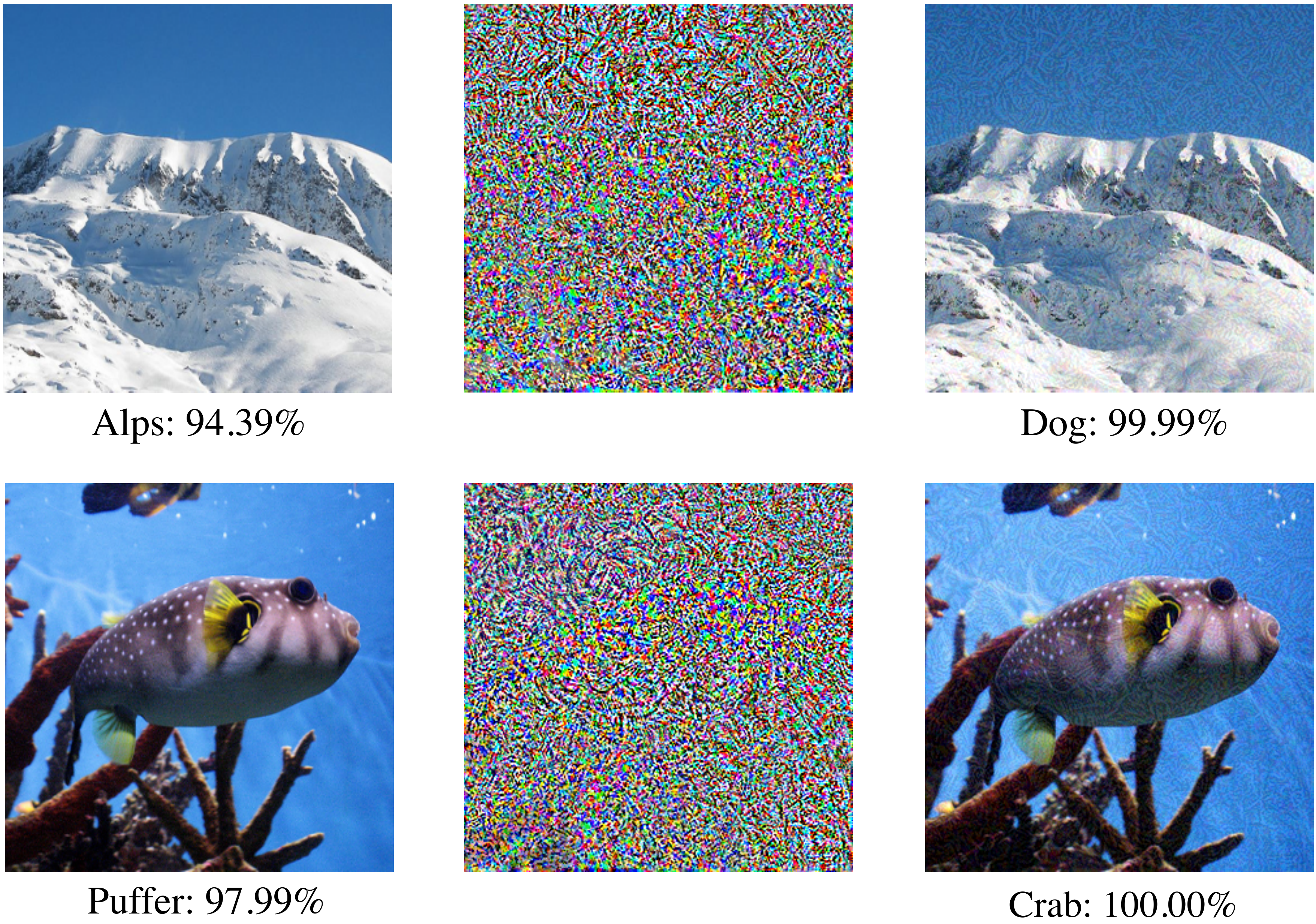}
    \caption{We show two adversarial examples generated by the proposed momentum iterative fast gradient sign method (MI-FGSM) for the Inception v3~\cite{Szegedy2015Rethinking} model.\textbf{ Left column}: the original images. \textbf{Middle column}: the adversarial noises by applying MI-FGSM for $10$ iterations. \textbf{Right column}: the generated adversarial images. We also show the predicted labels and probabilities of these images given by the Inception v3.}
    \vspace{-1ex}
    \label{fig:demo}
\end{figure}

Deep neural networks (DNNs) are challenged by their vulnerability to adversarial examples~\cite{szegedy2013intriguing,goodfellow2014explaining}, which are crafted by adding small, human-imperceptible noises to legitimate examples, but make a model output attacker-desired inaccurate predictions.
It has garnered an increasing attention to generating adversarial examples since it helps to identify the vulnerability of the models before they are launched. Besides, adversarial samples also facilitate various DNN algorithms to assess the robustness by providing more varied training data~\cite{goodfellow2014explaining,kurakin2017adversarial}.

With the knowledge of the structure and parameters of a given model, many methods can successfully generate adversarial examples in the white-box manner, including optimization-based methods such as box-constrained L-BFGS~\cite{szegedy2013intriguing}, one-step gradient-based methods such as fast gradient sign~\cite{goodfellow2014explaining} and iterative variants of gradient-based methods~\cite{kurakin2016adversarial}.
In general, a more severe issue of adversarial examples is their good transferability~\cite{szegedy2013intriguing,liu2016delving,Moosavi2016Universal}, \ie, the adversarial examples crafted for one model remain adversarial for others, thus making black-box attacks practical in real-world applications and posing real security issues.
The phenomenon of transferability is due to the fact that different machine learning models learn similar decision boundaries around a data point, making the adversarial examples crafted for one model also effective for others.


However, existing attack methods exhibit low efficacy when attacking black-box models, especially for those with a defense mechanism.
For example, ensemble adversarial training~\cite{tramer2017ensemble} significantly improves the robustness of deep neural networks and most of existing methods cannot successfully attack them in the black-box manner.
This fact largely attributes to the trade-off between the attack ability and the transferability.
In particular, the adversarial examples generated by optimization-based and iterative methods have poor transferability~\cite{kurakin2017adversarial}, and thus make black-box attacks less effective.
On the other hand, one-step gradient-based methods generate more transferable adversarial examples, however they usually have a low success rate for the white-box model~\cite{kurakin2017adversarial},
making it ineffective for black-box attacks.
Given the difficulties of practical black-box attacks, Papernot \etal~\cite{Papernot2016Practical} use adaptive queries to train a surrogate model to fully characterize the behavior of the target model and therefore turn the black-box attacks to white-box attacks.
However, it requires the full prediction confidences given by the target model and tremendous number of queries, especially for large scale datasets such as ImageNet~\cite{russakovsky2015imagenet}. Such requirements are impractical in real-world applications.
Therefore, we consider how to effectively attack a black-box model without knowing its architecture and parameters, and further, without querying.

In this paper, we propose a broad class of \textit{momentum iterative gradient-based} methods to boost the success rates of the generated adversarial examples.
Beyond iterative gradient-based methods that iteratively perturb the input with the gradients to maximize the loss function~\cite{goodfellow2014explaining}, momentum-based methods accumulate a velocity vector in the gradient direction of the loss function across iterations, for the purpose of stabilizing update directions and escaping from poor local maxima.
We show that the adversarial examples generated by momentum iterative methods have higher success rates in both white-box and black-box attacks.
The proposed methods alleviate the trade-off between the white-box attacks and the transferability, and act as a stronger attack algorithm than one-step methods~\cite{goodfellow2014explaining} and vanilla iterative methods~\cite{kurakin2016adversarial}.

To further improve the transferability of adversarial examples, we study several approaches for attacking an ensemble of models,
because if an adversarial example fools multiple models, it is more likely to remain adversarial for other black-box models~\cite{liu2016delving}.
We show that the adversarial examples generated by the momentum iterative methods for multiple models, can successfully fool robust models obtained by ensemble adversarial training~\cite{tramer2017ensemble} in the black-box manner. The findings in this paper raise new security issues for developing more robust deep learning models, with a hope that our attacks will be used as a benchmark to evaluate the robustness of various deep learning models and defense methods. 
In summary, we make the following contributions:
\begin{itemize}

\item We introduce a class of attack algorithms called momentum iterative gradient-based methods, in which we accumulate gradients of the loss function at each iteration to stabilize optimization and escape from poor local maxima.
\vspace{-1ex}
\item We study several ensemble approaches to attack multiple models simultaneously, which demonstrates a powerful capability of transferability by preserving a high success rate of attacks.
\vspace{-1ex}
\item We are the first to show that the models obtained by ensemble adversarial training with a powerful defense ability are also vulnerable to the black-box attacks.
\end{itemize}
\section{Backgrounds}

In this section, we provide the background knowledge as well as review the related works about adversarial attack and defense methods. Given a classifier $f(\bm{x}): \bm{x} \in \mathcal{X} \rightarrow y \in \mathcal{Y}$ that outputs a label $y$ as the prediction for an input $\bm{x}$, the goal of adversarial attacks is to seek an example $\bm{x}^*$ in the vicinity of $\bm{x}$ but is misclassified by the classifier. Specifically, there are two classes of adversarial examples---{\it non-targeted} and {\it targeted} ones. For a correctly classified input $\bm{x}$ with ground-truth label $y$ such that $f(\bm{x}) = y$, a non-targeted adversarial example $\bm{x}^*$ is crafted by adding small noise to $\bm{x}$ without changing the label, but misleads the classifier as $f(\bm{x}^*) \neq y$; and a targeted adversarial example aims to fool the classifier by outputting a specific label as $f(\bm{x}^*) = y^*$, where $y^*$ is the target label specified by the adversary, and $y^* \neq y$. In most cases, the $L_p$ norm of the adversarial noise is required to be less than an allowed value $\epsilon$ as $\|\bm{x}^* - \bm{x}\|_p \leq \epsilon$, where $p$ could be $0, 1, 2, \infty$.



\subsection{Attack methods}
\label{sec:attack}
Existing approaches for generating adversarial examples can be categorized into three groups. We introduce their non-targeted version of attacks here, and the targeted version can be simply derived.

\textbf{One-step gradient-based approaches}, such as the fast gradient sign method (FGSM)~\cite{goodfellow2014explaining}, find an adversarial example $\bm{x}^*$ by maximizing the loss function $J(\bm{x}^*,y)$, where $J$ is often the cross-entropy loss.
FGSM generates adversarial examples to meet the $L_{\infty}$ norm bound $\|\bm{x}^* - \bm{x}\|_{\infty} \leq \epsilon$ as
\begin{equation}
\bm{x}^* = \bm{x} + \epsilon\cdot\mathrm{sign}(\nabla_{\bm{x}}J(\bm{x},y)),
\label{eq:fgsm}
\end{equation}
where $\nabla_{\bm{x}}J(\bm{x},y)$ is the gradient of the loss function w.r.t. $\bm{x}$. 
The fast gradient method (FGM) is a generalization of FGSM to meet the $L_2$ norm bound $\|\bm{x}^* - \bm{x}\|_2 \leq \epsilon$ as
\begin{equation}
\bm{x}^* = \bm{x} + \epsilon\cdot \frac{\nabla_{\bm{x}}J(\bm{x},y)}{\|\nabla_{\bm{x}}J(\bm{x},y)\|_2}.
\label{eq:fg}
\end{equation}

\textbf{Iterative methods}~\cite{kurakin2016adversarial} iteratively apply fast gradient multiple times with a small step size $\alpha$. The iterative version of FGSM (I-FGSM) can be expressed as:
\begin{equation}
\bm{x}_0^* = \bm{x},\hspace{0.3cm} \bm{x}_{t+1}^* = \bm{x}_t^* + \alpha\cdot\mathrm{sign}(\nabla_{\bm{x}}J(\bm{x}_t^*,y)).
\label{eq:iter}
\end{equation}
To make the generated adversarial examples satisfy the $L_{\infty}$ (or $L_2$) bound, one can clip $\bm{x}_{t}^*$ into the $\epsilon$ vicinity of $\bm{x}$ or simply set $\alpha = \nicefrac{\epsilon}{T}$ with $T$ being the number of iterations.
It has been shown that iterative methods are stronger white-box adversaries than one-step methods at the cost of worse transferability~\cite{kurakin2017adversarial,tramer2017ensemble}.

\textbf{Optimization-based methods}~\cite{szegedy2013intriguing} directly optimize the distance between the real and adversarial examples subject to the misclassification of adversarial examples. Box-constrained L-BFGS can be used to solve such a problem. A more sophisticated way~\cite{Carlini2017Towards} is solving:
\begin{equation}
\argmin_{\bm{x}^*}\lambda\cdot\|\bm{x}^*-\bm{x}\|_p - J(\bm{x}^*, y).
\end{equation}
Since it directly optimizes the distance between an adversarial example and the corresponding real example, there is no guarantee that the $L_{\infty}$ ($L_2$) distance is less than the required value.
Optimization-based methods also lack the efficacy in black-box attacks just like iterative methods.

\subsection{Defense methods}


Among many attempts~\cite{Metzen2017On,Dong2017Towards,Pang2017Robust,kurakin2017adversarial,tramer2017ensemble,Papernot2016Distillation,Li2017Dropout}, adversarial training is the most extensively investigated way to increase the robustness of DNNs~\cite{goodfellow2014explaining,kurakin2017adversarial,tramer2017ensemble}. By injecting adversarial examples into the training procedure, 
the adversarially trained models learn to resist the perturbations in the gradient direction of the loss function.
However, they do not confer robustness to black-box attacks due to the coupling of the generation of adversarial examples and the parameters being trained.
Ensemble adversarial training~\cite{tramer2017ensemble} augments the training data with the adversarial samples produced not only from the model being trained, but also from other hold-out models. Therefore, the ensemble adversarially trained models are robust against one-step attacks and black-box attacks.

\section{Methodology}
\label{sec:momentum}
In this paper, we propose a broad class of \textbf{momentum iterative gradient-based methods} to generate adversarial examples, which can fool white-box models as well as black-box models.
In this section, we elaborate the proposed algorithms. We first illustrate how to integrate momentum into iterative FGSM, which induces a momentum iterative fast gradient sign method (MI-FGSM) to generate adversarial examples satisfying the $L_{\infty}$ norm restriction in the non-targeted attack fashion.
We then present several methods on how to efficiently attack an ensemble of models.
Finally, we extend MI-FGSM to $L_2$ norm bound and targeted attacks, yielding a broad class of attack methods.

\subsection{Momentum iterative fast gradient sign method}
The momentum method~\cite{Poljak1964Some} is a technique for accelerating gradient descent algorithms by accumulating a velocity vector in the gradient direction of the loss function across iterations. The memorization of previous gradients helps to barrel through narrow valleys, small humps and poor local minima or maxima~\cite{Korczak1998Optimization}.
The momentum method also shows its effectiveness in stochastic gradient descent to stabilize the updates~\cite{Sutskever2013On}. We apply the idea of momentum to generate adversarial examples and obtain tremendous benefits.

To generate a non-targeted adversarial example $\bm{x}^*$ from a real example $\bm{x}$, which satisfies the $L_{\infty}$ norm bound, gradient-based approaches seek the adversarial example by solving the constrained optimization problem
\begin{equation}
\argmax_{\bm{x}^*} J(\bm{x}^*, y), \hspace{0.4cm} \mathrm{s.t.} \hspace{0.2cm} \|\bm{x}^* - \bm{x}\|_{\infty} \leq \epsilon,
\end{equation}
where $\epsilon$ is the size of adversarial perturbation.
FGSM generates an adversarial example by applying the sign of the gradient to a real example only once (in Eq.~\eqref{eq:fgsm}) by the assumption of linearity of the decision boundary around the data point.
However in practice, the linear assumption may not hold when the distortion is large~\cite{liu2016delving}, which makes the adversarial example generated by FGSM  ``underfits'' the model, limiting its attack ability. 
In contrast, iterative FGSM greedily moves the adversarial example in the direction of the sign of the gradient in each iteration (in Eq.~\eqref{eq:iter}). Therefore, the adversarial example can easily drop into poor local maxima and ``overfit'' the model, which is not likely to transfer across models.

\begin{algorithm}[t]
\small
\caption{MI-FGSM}
\label{alg:MI-FGSM}
\begin{algorithmic}[1]
\Require A classifier $f$ with loss function $J$; a real example $\bm{x}$ and ground-truth label $y$;
\Require The size of perturbation $\epsilon$; iterations $T$ and decay factor $\mu$.
\Ensure
An adversarial example $\bm{x}^*$ with $\|\bm{x}^* - \bm{x}\|_{\infty} \leq \epsilon$.
\State $\alpha = \nicefrac{\epsilon}{T}$;
\State $\bm{g}_0 = 0$; $\bm{x}_0^* = \bm{x}$;
\For {$t = 0$ to $T-1$}
\State Input $\bm{x}_t^*$ to $f$ and obtain the gradient $\nabla_{\bm{x}}J(\bm{x}_t^*,y)$;
\State Update $\bm{g}_{t+1}$ by accumulating the velocity vector in the gradient direction as 
\vspace{-2ex}
\begin{equation}
\label{eq:momentum}
\bm{g}_{t+1} = \mu \cdot \bm{g}_{t} + \frac{\nabla_{\bm{x}}J(\bm{x}_{t}^*,y)}{\|\nabla_{\bm{x}}J(\bm{x}_{t}^*,y)\|_1};
\end{equation}
\vspace{-3ex}
\State Update $\bm{x}_{t+1}^*$ by applying the sign gradient as 
\vspace{-1.5ex}
\begin{equation}
\label{eq:update}
\bm{x}_{t+1}^* = \bm{x}_{t}^* + \alpha\cdot\mathrm{sign}(\bm{g}_{t+1});
\end{equation}
\vspace{-4ex}
\EndFor
\Return $\bm{x}^* = \bm{x}_T^*$.
\end{algorithmic}
\end{algorithm}

In order to break such a dilemma, we integrate momentum into the iterative FGSM for the purpose of stabilizing update directions and escaping from poor local maxima. Therefore, the momentum-based method remains the transferability of adversarial examples when increasing iterations, and at the same time acts as a strong adversary for the white-box models like iterative FGSM. It alleviates the trade-off between the attack ability and the transferability, demonstrating strong black-box attacks.

The momentum iterative fast gradient sign method (MI-FGSM) is summarized in Algorithm~\ref{alg:MI-FGSM}.
Specifically, $\bm{g}_t$ gathers the gradients of the first $t$ iterations with a decay factor $\mu$, defined in Eq.~\eqref{eq:momentum}. Then the adversarial example $\bm{x}_t^*$ until the $t$-th iteration is perturbed in the direction of the sign of $\bm{g}_t$ with a step size $\alpha$ in Eq.~\eqref{eq:update}. If $\mu$ equals to $0$, MI-FGSM degenerates to the iterative FGSM.
In each iteration, the current gradient $\nabla_{\bm{x}}J(\bm{x}_t^*,y)$ is normalized by the $L_1$ distance (any distance measure is feasible) of itself, because we notice that the scale of the gradients in different iterations varies in magnitude.

\subsection{Attacking ensemble of models}
\label{sec:ensemble}

In this section, we study how to attack an ensemble of models efficiently. Ensemble methods have been broadly adopted in researches and competitions for enhancing the performance and improving the robustness~\cite{Hansen1990Neural,Krogh1994Neural,Caruana2004Ensemble}.
The idea of ensemble can also be applied to adversarial attacks, due to the fact that if an example remains adversarial for multiple models, it may capture an intrinsic direction that always fools these models and is more likely to transfer to other models at the same time~\cite{liu2016delving}, thus enabling powerful black-box attacks.

We propose to attack multiple models whose \textit{logit} activations\footnote{Logits are the input values to softmax.} are fused together, and we call this method \textit{ensemble in logits}.
Because the logits capture the logarithm relationships between the probability predictions, an ensemble of models fused by logits aggregates the fine detailed outputs of all models, whose vulnerability can be easily discovered.
Specifically, to attack an ensemble of $K$ models, we fuse the logits as 
\begin{equation}
\bm{l}(\bm{x}) = \textstyle\sum_{k=1}^K w_k \bm{l}_k(\bm{x}),
\end{equation}
where $\bm{l}_k(\bm{x})$ are the logits of the $k$-th model, $w_k$ is the ensemble weight with $w_k \geq 0$ and $\sum_{k=1}^K w_k = 1$.
The loss function $J(\bm{x}, y)$ is defined as the softmax cross-entropy loss given the ground-truth label $y$ and the logits $\bm{l}(\bm{x})$
\begin{equation}\label{equ_cross_entropy}
J(\bm{x}, y) = -\mathbf{1}_{y}\cdot\log(\mathrm{softmax}(\bm{l}(\bm{x}))),
\end{equation}
where $\mathbf{1}_{y}$ is the one-hot encoding of $y$.
We summarize the MI-FGSM algorithm for attacking multiple models whose logits are averaged in Algorithm~\ref{alg:MI-FGSM-multi}.
\begin{algorithm}[t]
\small
\caption{MI-FGSM for an ensemble of models}
\label{alg:MI-FGSM-multi}
\begin{algorithmic}[1]
\Require The logits of $K$ classifiers $\bm{l}_1, \bm{l}_2, ..., \bm{l}_K$; ensemble weights $w_1, w_2, ..., w_K$; a real example $\bm{x}$ and ground-truth label $y$;
\Require The size of perturbation $\epsilon$; iterations $T$ and decay factor $\mu$.
\Ensure
An adversarial example $\bm{x}^*$ with $\|\bm{x}^* - \bm{x}\|_{\infty} \leq \epsilon$.
\State $\alpha = \nicefrac{\epsilon}{T}$;
\State $\bm{g}_0 = 0$; $\bm{x}_0^* = \bm{x}$;
\For {$t = 0$ to $T-1$}
\State Input $\bm{x}_t^*$ and output $\bm{l}_k(\bm{x}_t^*)$ for $k = 1, 2, ..., K$;
\State Fuse the logits as $\bm{l}(\bm{x}_t^*) = \sum_{k=1}^Kw_k\bm{l}_k(\bm{x}_t^*)$;
\State Get softmax cross-entropy loss $J(\bm{x}_t^*,y)$ based on $\bm{l}(\bm{x}_t^*)$ and Eq.~\eqref{equ_cross_entropy};
\State Obtain the gradient $\nabla_{\bm{x}}J(\bm{x}_t^*,y)$;
\State Update $\bm{g}_{t+1}$ by Eq.~\eqref{eq:momentum};
\State Update $\bm{x}_{t+1}^*$ by Eq.~\eqref{eq:update};
\EndFor
\Return $\bm{x}^* = \bm{x}_T^*$.
\end{algorithmic}
\end{algorithm}

For comparison, we also introduce two alternative ensemble schemes, one of which is already studied~\cite{liu2016delving}.
Specifically, $K$ models can be averaged in predictions~\cite{liu2016delving} as $\bm{p}(\bm{x}) = \sum_{k=1}^K w_k \bm{p}_k(\bm{x})$, where $\bm{p}_k(\bm{x})$ is the predicted probability of the $k$-th model given input $\bm{x}$. $K$ models can also be averaged in loss as $J(\bm{x}, y) = \sum_{k=1}^K w_k J_k(\bm{x}, y)$.
In these three methods, the only difference is where to combine the outputs of multiple models, but they result in different attack abilities. We empirically find that the ensemble in logits performs better than the ensemble in predictions and the ensemble in loss, among various attack methods and various models in the ensemble, which will be demonstrated in Sec.~\ref{sec:exp-ensemble}.

\begin{table*}
\footnotesize
\begin{center}
\begin{tabular}{c|p{10ex}<{\centering}|p{10ex}<{\centering}|p{10ex}<{\centering}|p{10ex}<{\centering}|p{10ex}<{\centering}|p{10ex}<{\centering}|p{10ex}<{\centering}|c}

\hline
& Attack & Inc-v3 & Inc-v4 & IncRes-v2 & Res-152 & Inc-v3\textsubscript{ens3} & Inc-v3\textsubscript{ens4} & IncRes-v2\textsubscript{ens} \\
\hline\hline

\multirow{3}{*}{Inc-v3} & FGSM & 72.3$^*$ & 28.2 & 26.2 & 25.3 & 11.3 & 10.9 & 4.8\\
& I-FGSM & \textbf{100.0}$^*$ & 22.8 & 19.9 & 16.2 & 7.5 & 6.4 & 4.1\\
& MI-FGSM & \textbf{100.0}$^*$ & \textbf{48.8} & \textbf{48.0} & \textbf{35.6} & \textbf{15.1} & \textbf{15.2} & \textbf{7.8}\\
\hline
\multirow{3}{*}{Inc-v4} & FGSM & 32.7 & 61.0$^*$ & 26.6 & 27.2 & 13.7 & 11.9 & 6.2 \\
& I-FGSM & 35.8 & \textbf{99.9}$^*$ & 24.7 & 19.3 & 7.8 & 6.8 & 4.9 \\
& MI-FGSM & \textbf{65.6} & \textbf{99.9}$^*$ & \textbf{54.9} & \textbf{46.3} & \textbf{19.8} & \textbf{17.4} & \textbf{9.6} \\
\hline
\multirow{3}{*}{IncRes-v2} & FGSM & 32.6 & 28.1 & 55.3$^*$ & 25.8 & 13.1 & 12.1 & 7.5 \\
& I-FGSM & 37.8 & 20.8 & \textbf{99.6}$^*$ & 22.8 & 8.9 & 7.8 & 5.8 \\
& MI-FGSM & \textbf{69.8} & \textbf{62.1} & 99.5$^*$ & \textbf{50.6} & \textbf{26.1} & \textbf{20.9} & \textbf{15.7} \\
\hline
\multirow{3}{*}{Res-152} & FGSM & 35.0 & 28.2 & 27.5 & 72.9$^*$ & 14.6 & 13.2 & 7.5\\
& I-FGSM & 26.7 & 22.7 & 21.2 & \textbf{98.6}$^*$ & 9.3 & 8.9 & 6.2 \\
& MI-FGSM & \textbf{53.6} & \textbf{48.9} & \textbf{44.7} & 98.5$^*$ & \textbf{22.1} & \textbf{21.7} & \textbf{12.9} \\

\hline
\end{tabular}
\end{center}
\caption{The success rates (\%) of non-targeted adversarial attacks against seven models we study. The adversarial examples are crafted for Inc-v3, Inc-v4, IncRes-v2 and Res-152 respectively using FGSM, I-FGSM and MI-FGSM. $^*$ indicates the white-box attacks.}
\label{tab:single-model}
\vspace{-2ex}
\end{table*}

\subsection{Extensions}

The momentum iterative methods can be easily generalized to other attack settings. By replacing the current gradient with the accumulated gradient of all previous steps, any iterative method can be extended to its momentum variant.
Here we introduce the methods for generating adversarial examples in terms of the $L_2$ norm bound attacks and the targeted attacks.

To find an adversarial examples within the $\epsilon$ vicinity of a real example measured by $L_2$ distance as $\|\bm{x}^* - \bm{x}\|_2 \leq \epsilon$, the momentum variant of iterative fast gradient method (MI-FGM) can be written as
\begin{equation}
\bm{x}_{t+1}^* = \bm{x}_t^* + \alpha\cdot\frac{\bm{g}_{t+1}}{\|\bm{g}_{t+1}\|_2},
\label{eq:mi-fgs}
\end{equation}
where $\bm{g}_{t+1}$ is defined in Eq.~\eqref{eq:momentum} and $\alpha=\nicefrac{\epsilon}{T}$ with $T$ standing for the total number of iterations.

For targeted attacks, the objective for finding an adversarial example misclassified as a target class $y^*$ is to minimize the loss function $J(\bm{x}^*, y^*)$.
The accumulated gradient is derived as
\begin{equation}
\bm{g}_{t+1} = \mu \cdot \bm{g}_t + \frac{J(\bm{x}_t^*,y^*)}{\|\nabla_{\bm{x}}J(\bm{x}_t^*,y^*)\|_1}.
\end{equation}
The targeted MI-FGSM with an $L_{\infty}$ norm bound is
\begin{equation}
\bm{x}_{t+1}^* = \bm{x}_t^* - \alpha\cdot\mathrm{sign}(\bm{g}_{t+1}),
\end{equation}
and the targeted MI-FGM with an $L_2$ norm bound is
\begin{equation}
\bm{x}_{t+1}^* = \bm{x}_t^* - \alpha\cdot\frac{\bm{g}_{t+1}}{\|\bm{g}_{t+1}\|_2}.
\end{equation}
Therefore, we introduce a broad class of momentum iterative methods for attacks in various settings, whose effectiveness is demonstrated in Sec.~\ref{sec:exp}.

\section{Experiments}
\label{sec:exp}
In this section, we conduct extensive experiments on the ImageNet dataset~\cite{russakovsky2015imagenet} to validate the effectiveness of the proposed methods. We first specify the experimental settings in Sec.~\ref{sec:setup}. Then we report the results for attacking a single model in Sec.~\ref{sec:attack-single-model} and an ensemble of models in Sec.~\ref{sec:ensemble-exp}. 
Our methods won both the NIPS 2017 Non-targeted and Targeted Adversarial Attack competitions, with the configurations introduced in Sec.~\ref{sec:competition}.

\subsection{Setup}

\label{sec:setup}
 We study seven models, four of which are normally trained models---Inception v3 (Inc-v3)~\cite{Szegedy2015Rethinking}, Inception v4 (Inc-v4), Inception Resnet v2 (IncRes-v2)~\cite{szegedy2017inception}, Resnet v2-152 (Res-152)~\cite{he2016identity} and the other three of which are trained by ensemble adversarial training---Inc-v3\textsubscript{ens3}, Inc-v3\textsubscript{ens4}, IncRes-v2\textsubscript{ens}~\cite{tramer2017ensemble}. We will simply call the last three models as ``adversarially trained models'' without ambiguity.

It is less meaningful to study the success rates of attacks if the models cannot classify the original image correctly. Therefore, we randomly choose $1000$ images belonging to the $1000$ categories from the ILSVRC 2012 validation set, which are all correctly classified by them.

In our experiments, we compare our methods to one-step gradient-based methods and iterative methods. Since optimization-based methods cannot explicitly control the distance between the adversarial examples and the corresponding real examples, they are not directly comparable to ours, but they have similar properties with iterative methods as discussed in Sec.~\ref{sec:attack}.
For clarity, we only report the results based on $L_{\infty}$ norm bound for non-targeted attacks, and leave the results based on $L_2$ norm bound and targeted attacks in the supplementary material. The findings in this paper are general across different attack settings.

\subsection{Attacking a single model}
\label{sec:attack-single-model}

We report in Table~\ref{tab:single-model} the success rates of attacks against the models we consider. The adversarial examples are generated for Inc-v3, Inc-v4, InvRes-v2 and Res-152 respectively using FGSM, iterative FGSM (I-FGSM) and MI-FGSM attack methods.
The success rates are the misclassification rates of the corresponding models with adversarial images as inputs.
The maximum perturbation $\epsilon$ is set to $16$ among all experiments, with pixel value in $[0,255]$. The number of iterations is $10$ for I-FGSM and MI-FGSM, and the decay factor $\mu$ is $1.0$, which will be studied in Sec.~\ref{sec:mu}.

From the table, we can observe that MI-FGSM remains as a strong white-box adversary like I-FGSM since it can attack a white-box model with a near $100\%$ success rate.
On the other hand, it can be seen that I-FGSM reduces the success rates for black-box attacks than one-step FGSM. But by integrating momentum, our MI-FGSM outperforms both FGSM and I-FGSM in black-box attacks significantly. It obtains more than $2$ times of the success rates than I-FGSM in most black-box attack cases, demonstrating the effectiveness of the proposed algorithm.
We show two adversarial images in Fig.~\ref{fig:demo} generated for Inc-v3.

It should be noted that although our method greatly improves the success rates for black-box attacks, it is still ineffective for attacking adversarially trained models (\eg, less than $16\%$ for IncRes-v2\textsubscript{ens}) in the black-box manner. Later we show that ensemble-based approaches greatly improve the results in Sec.~\ref{sec:ensemble-exp}.
Next, we study several aspects of MI-FGSM that are different from vanilla iterative methods, to further explain why it performs well in practice.

\vspace{-2ex}
\subsubsection{Decay factor $\mu$}
\label{sec:mu}

\begin{figure}[!t]
  \centering
    \includegraphics[width=0.85\linewidth]{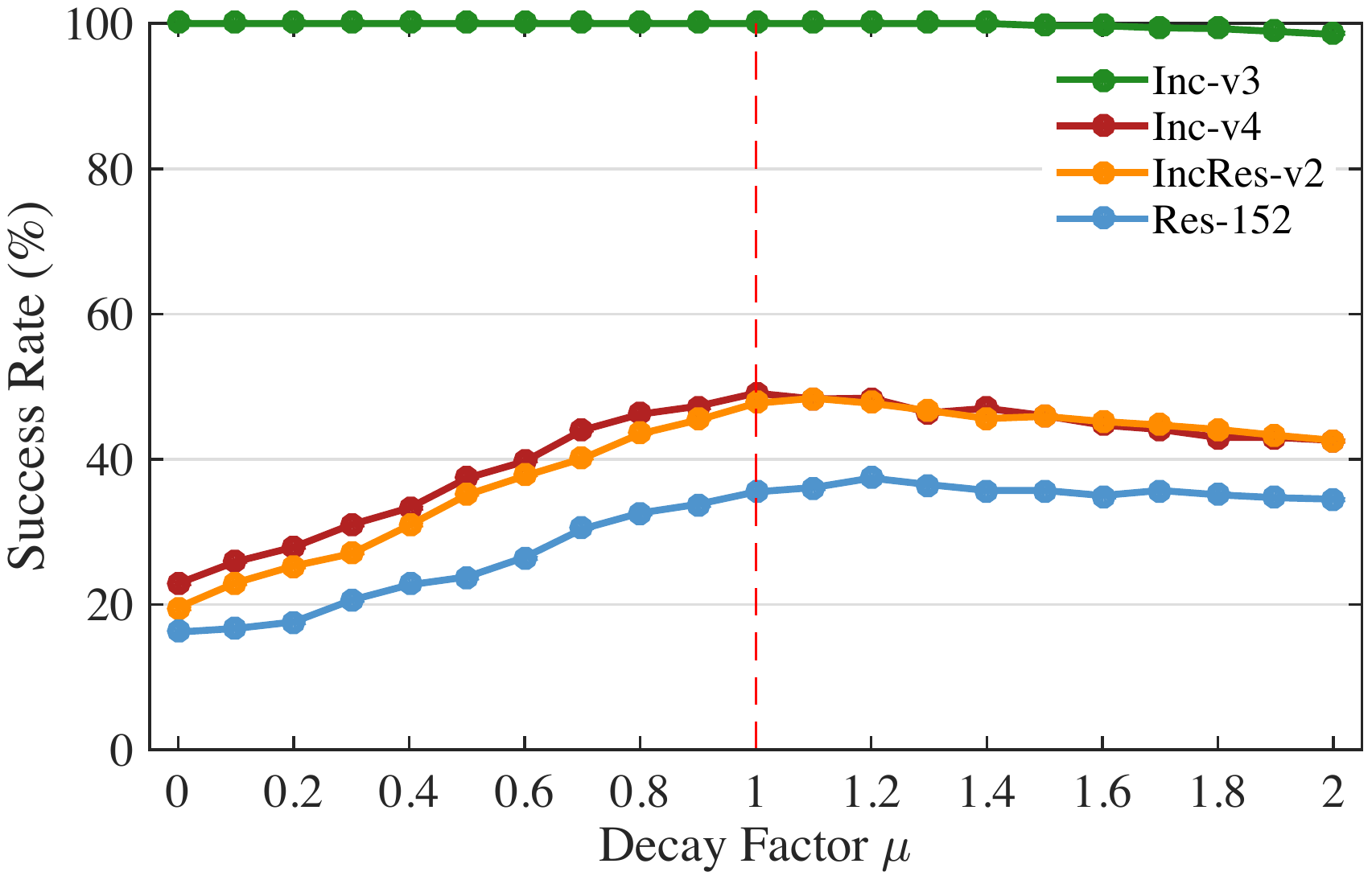}
    \caption{The success rates (\%) of the adversarial examples generated for Inc-v3 against Inc-v3 (white-box), Inc-v4, IncRes-v2 and Res-152 (black-box), with $\mu$ ranging from $0.0$ to $2.0$.}
    \label{fig:momentum}
    \vspace{-2ex}
\end{figure}

The decay factor $\mu$ plays a key role for improving the success rates of attacks. If $\mu = 0 $, momentum-based iterative methods trivially turn to vanilla iterative methods. Therefore, we study the appropriate value of the decay factor.
We attack Inc-v3 model by MI-FGSM with the perturbation $\epsilon=16$, the number of iterations $10$, and the decay factor ranging from $0.0$ to $2.0$ with a granularity $0.1$. We show the success rates of the generated adversarial examples against Inc-v3, Inc-v4, IncRes-v2 and Res-152 in Fig.~\ref{fig:momentum}. The curve of the success rate against a black-box model is unimodal whose maximum value is obtained at around $\mu = 1.0$. When $\mu = 1.0$, another interpretation of $\bm{g}_t$ defined in Eq.~\eqref{eq:momentum} is that it simply adds up all previous gradients to perform the current update.

\vspace{-2ex}
\subsubsection{The number of iterations}

\begin{figure}[!t]
  \centering
    \includegraphics[width=0.85\linewidth]{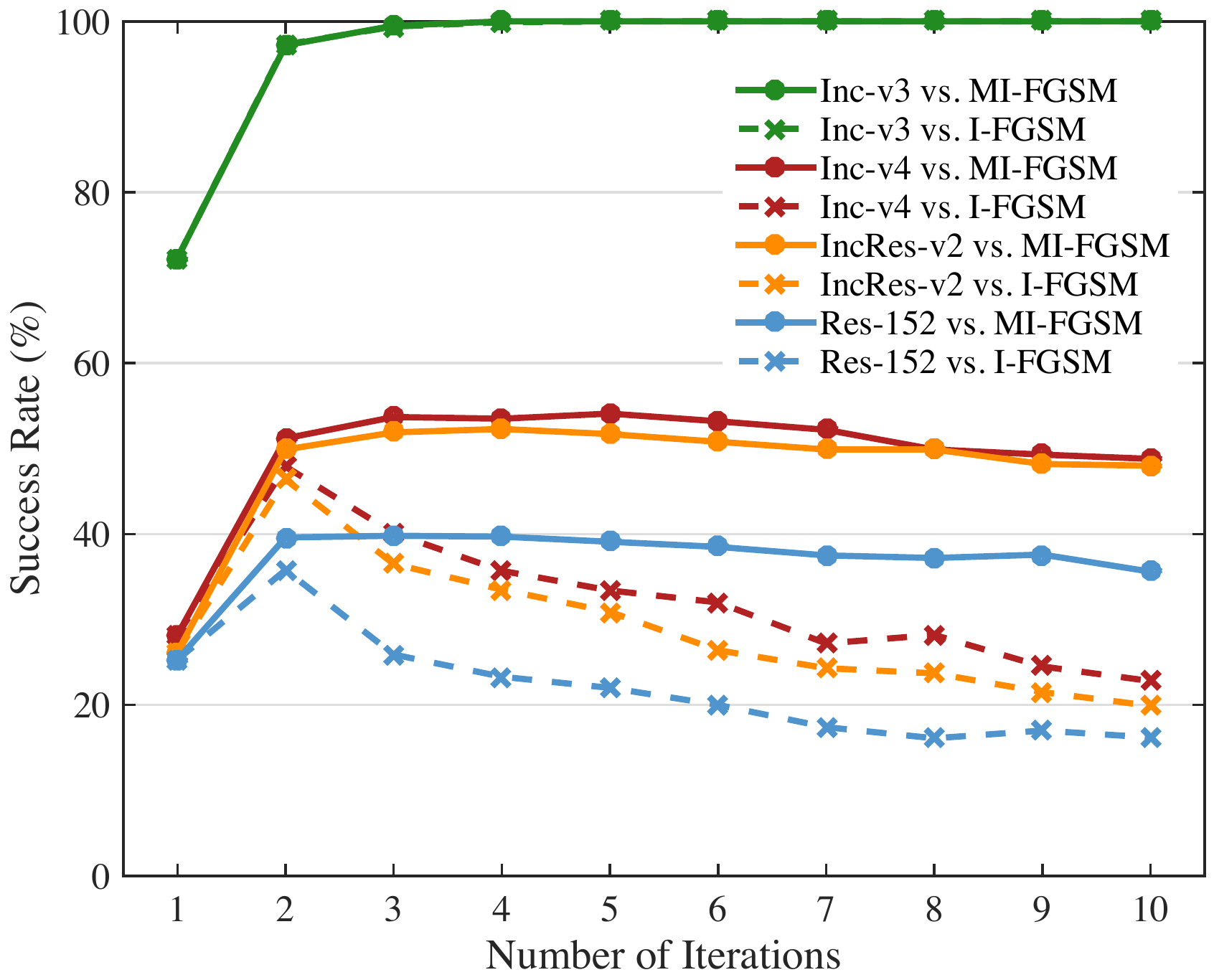}
    \caption{The success rates (\%) of the adversarial examples generated for Inc-v3 model against Inc-v3 (white-box), Inc-v4, IncRes-v2 and Res-152 (black-box). We compare the results of I-FGSM and MI-FGSM with different iterations. Please note that the curves of Inc-v3 vs. MI-FGSM and Inc-v3 vs. I-FGSM overlap together.}
    \label{fig:iter}
    \vspace{-2ex}
\end{figure}

We then study the effect of the number of iterations on the success rates when using I-FGSM and MI-FGSM. We adopt the same hyper-parameters (\ie, $\epsilon = 16$, $\mu = 1.0$) for attacking Inc-v3 model with the number of iterations ranging from $1$ to $10$, and then evaluate the success rates of adversarial examples against Inc-v3, Inc-v4, IncRes-v2 and Res-152 models, with the results shown in Fig.~\ref{fig:iter}.

It can be observed that when increasing the number of iterations, the success rate of I-FGSM against a black-box model gradually decreases, while that of MI-FGSM maintains at a high value.
The results prove our argument that the adversarial examples generated by iterative methods easily overfit a white-box model and are not likely to transfer across models. But momentum-based iterative methods help to alleviate the trade-off between the white-box attacks and the transferability, thus demonstrating a strong attack ability for white-box and black-box models simultaneously.

\begin{table*}
\footnotesize
\begin{center}
\begin{tabular}{c|c|p{10ex}<{\centering}|p{10ex}<{\centering}|p{10ex}<{\centering}|p{10ex}<{\centering}|p{10ex}<{\centering}|p{10ex}<{\centering}}

\hline
\multirow{2}{*}{} & \multirow{2}{*}{Ensemble method} & \multicolumn{2}{c|}{FGSM} & \multicolumn{2}{c|}{I-FGSM} & \multicolumn{2}{c}{MI-FGSM} \\
\cline{3-8}
& & Ensemble & Hold-out & Ensemble & Hold-out & Ensemble & Hold-out \\
\hline\hline

\multirow{3}{*}{-Inc-v3} & Logits & \textbf{55.7} & \textbf{45.7} & \textbf{99.7} & \textbf{72.1} & \textbf{99.6} & \textbf{87.9}\\
& Predictions & 52.3 & 42.7 & 95.1 & 62.7 & 97.1 & 83.3\\
& Loss & 50.5 & 42.2 & 93.8 & 63.1 & 97.0 & 81.9\\
\hline
\multirow{3}{*}{-Inc-v4} & Logits & \textbf{56.1} & \textbf{39.9} & \textbf{99.8} & \textbf{61.0} & \textbf{99.5} & \textbf{81.2} \\
& Predictions & 50.9 & 36.5 & 95.5 & 52.4 & 97.1 & 77.4 \\
& Loss & 49.3 & 36.2 & 93.9 & 50.2 & 96.1 & 72.5 \\
\hline
\multirow{3}{*}{-IncRes-v2} & Logits & \textbf{57.2} & \textbf{38.8} & \textbf{99.5} & \textbf{54.4} & \textbf{99.5} & \textbf{76.5} \\
& Predictions & 52.1 & 35.8 & 97.1 & 46.9 & 98.0 & 73.9 \\
& Loss & 50.7 & 35.2 & 96.2 & 45.9 & 97.4 & 70.8 \\
\hline
\multirow{3}{*}{-Res-152} & Logits & \textbf{53.5} & \textbf{35.9} & 99.6 & \textbf{43.5} & 99.6 & \textbf{69.6} \\
& Predictions & 51.9 & 34.6 & \textbf{99.9} & 41.0 & \textbf{99.8} & 67.0 \\
& Loss & 50.4 & 34.1 & 98.2 & 40.1 & 98.8 & 65.2 \\
\hline
\end{tabular}
\end{center}
\vspace{-2ex}
\caption{The success rates (\%) of non-targeted adversarial attacks of three ensemble methods. We report the results for an ensemble of white-box models and a hold-out black-box target model. We study four models---Inc-v3, Inc-v4, IncRes-v2 and Res-152. In each row, ``-'' indicates the name of the hold-out model and the adversarial examples are generated for the ensemble of the other three models by FGSM, I-FGSM and MI-FGSM respectively. Ensemble in logits consistently outperform other methods.}
\label{tab:ensemble-methods}
\vspace{-2ex}
\end{table*}

\vspace{-2ex}
\subsubsection{Update directions}
\begin{figure}[!t]
  \centering
    \includegraphics[width=0.85\linewidth]{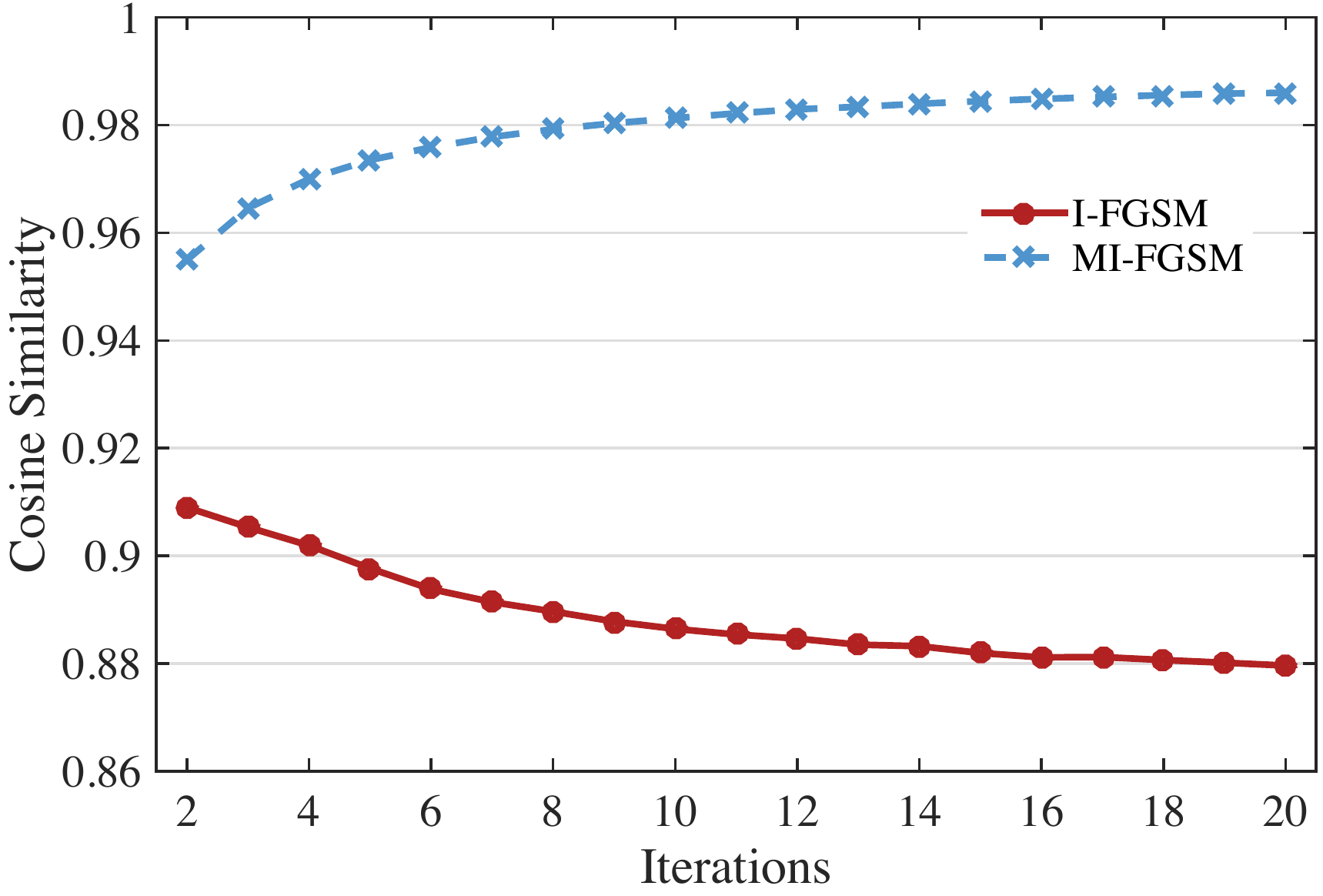}
    \caption{The cosine similarity of two successive perturbations in I-FGSM and MI-FGSM when attacking Inc-v3 model. The results are averaged over $1000$ images.}
    \label{fig:cos}
    \vspace{-2ex}
\end{figure}

To interpret why MI-FGSM demonstrates better transferability, we further examine the update directions given by I-FGSM and MI-FGSM along the iterations.
We calculate the cosine similarity of two successive perturbations and show the results in Fig.~\ref{fig:cos} when attacking Inc-v3. The update direction of MI-FGSM is more stable than that of I-FGSM due to the larger value of cosine similarity in MI-FGSM.

Recall that the transferability comes from the fact that models learn similar decision boundaries around a data point~\cite{liu2016delving}.
Although the decision boundaries are similar, they are unlikely the same due to the highly non-linear structure of DNNs.
So there may exist some exceptional decision regions around a data point for a model (holes as shown in Fig.~4\&5 in~\cite{liu2016delving}), which are hard to transfer to other models.
These regions correspond to poor local maxima in the optimization process and the iterative methods can easily trap into such regions, resulting in less transferable adversarial examples.
On the other hand, the stabilized update directions obtained by the momentum methods as observed in Fig.~\ref{fig:cos} can help to escape from these exceptional regions, resulting in better transferability for adversarial attacks.
Another interpretation is that the stabilized updated directions make the $L_2$ norm of the perturbations larger, which may be helpful for the transferability.

\vspace{-2ex}
\subsubsection{The size of perturbation}
\begin{figure}[!t]
  \centering
    \includegraphics[width=0.85\linewidth]{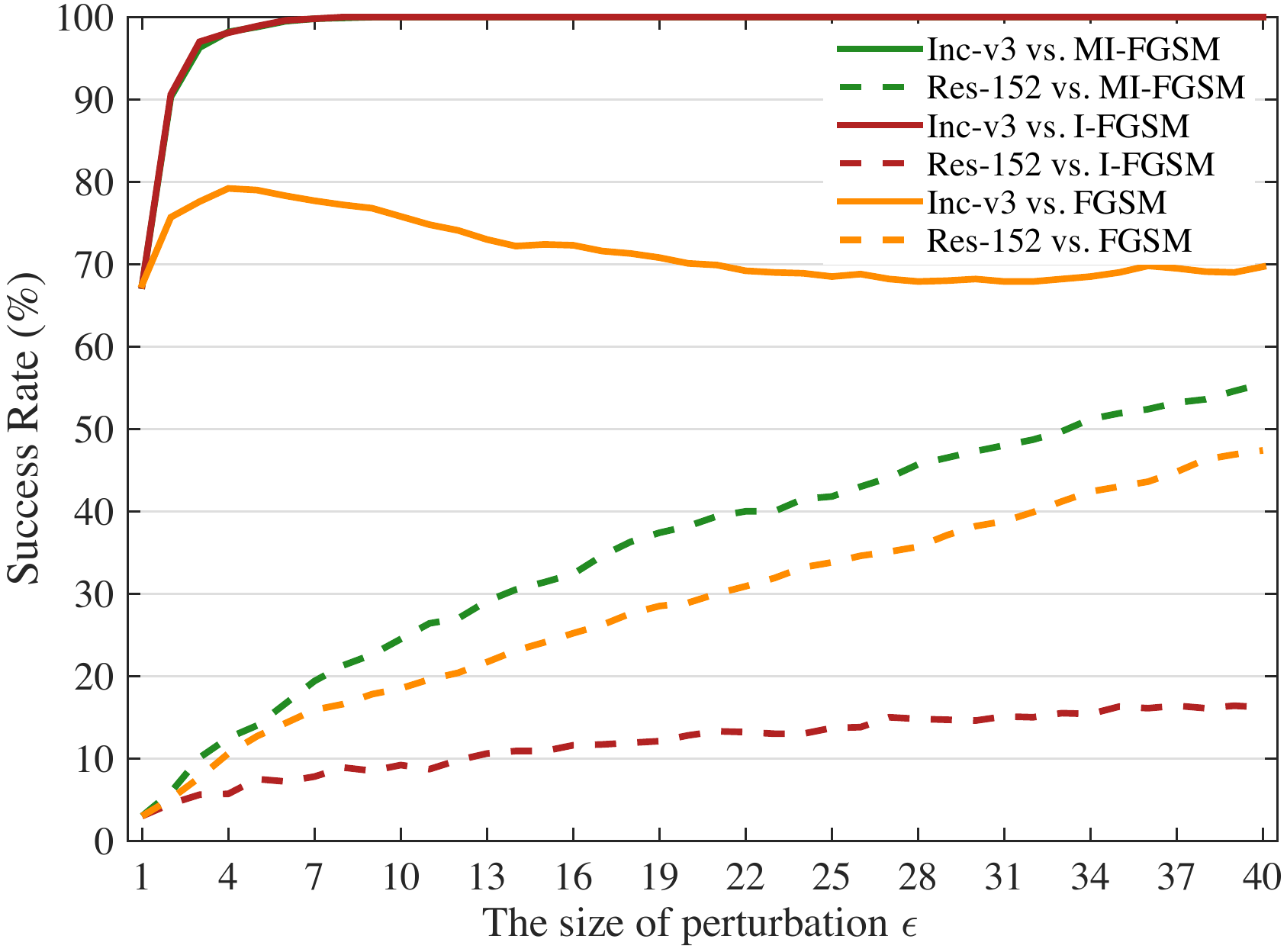}
    \caption{The success rates (\%) of the adversarial examples generated for Inc-v3 against Inc-v3 (white-box) and Res-152 (black-box). We compare the results of FGSM, I-FGSM and MI-FGSM with different size of perturbation. The curves of Inc-v3 vs. MI-FGSM and Inc-v3 vs. I-FGSM overlap together.}
    \label{fig:size}
    \vspace{-2ex}
\end{figure}

We finally study the influence of the size of adversarial perturbation on the success rates.
We attack Inc-v3 model by FGSM, I-FGSM and MI-FGSM with $\epsilon$ ranging from $1$ to $40$ with the image intensity  $[0,255]$, and evaluate the performance on the white-box model Inc-v3 and a black-box model Res-152. In our experiments, we set the step size $\alpha$ in I-FGSM and MI-FGSM to $1$, so the number of iterations grows linearly with the size of perturbation $\epsilon$.
The results are shown in Fig.~\ref{fig:size}.

For the white-box attack, iterative methods reach the $100\%$ success rate soon, but the success rate of one-step FGSM decreases when the perturbation is large. The phenomenon largely attributes to the inappropriate assumption of the linearity of the decision boundary when the perturbation is large~\cite{liu2016delving}.
For the black-box attacks, although the success rates of these three methods grow linearly with the size of perturbation, MI-FGSM's success rate grows faster. 
In other words, to attack a black-box model with a required success rate, MI-FGSM can use a smaller perturbation, which is more visually indistinguishable for humans.

\subsection{Attacking an ensemble of models}
\label{sec:ensemble-exp}
In this section, we show the experimental results of attacking an ensemble of models. We first compare the three ensemble methods introduced in Sec.~\ref{sec:ensemble}, and then demonstrate that the adversarially trained models are vulnerable to our black-box attacks. 

\vspace{-2ex}
\subsubsection{Comparison of ensemble methods}
\label{sec:exp-ensemble}

We compare the ensemble methods for attacks in this section. We include four models in our study, which are Inc-v3, Inc-v4, IncRes-v2 and Res-152. In our experiments, we keep one model as the hold-out black-box model and attack an ensemble of the other three models by FGSM, I-FGSM and MI-FGSM respectively, to fully compare the results of the three ensemble methods, \ie, ensemble in logits, ensemble in predictions and ensemble in loss.
We set $\epsilon$ to $16$, the number of iterations in I-FGSM and MI-FGSM to $10$, $\mu$ in MI-FGSM to $1.0$, and the ensemble weights equally. The results are shown in Table~\ref{tab:ensemble-methods}.

It can be observed that the ensemble in logits outperforms the ensemble in predictions and the ensemble in loss consistently among all the attack methods and different models in the ensemble for both the white-box and black-box attacks. Therefore, the ensemble in logits scheme is more suitable for adversarial attacks.

Another observation from Table~\ref{tab:ensemble-methods} is that the adversarial examples generated by MI-FGSM transfer at a high rate, enabling strong black-box attacks. For example, by attacking an ensemble of Inc-v4, IncRes-v2 and Res-152 fused in logits without Inc-v3, the generated adversarial examples can fool Inc-v3 with a $87.9\%$ success rate. Normally trained models show their great vulnerability against such an attack.

\vspace{-2ex}
\subsubsection{Attacking adversarially trained models}
\label{sec:ensemble-non-targeted}


\begin{table}
\footnotesize
\begin{center}
\begin{tabular}{c|p{10ex}<{\centering}|p{10ex}<{\centering}|p{10ex}<{\centering}}

\hline
& Attack & Ensemble & Hold-out \\
\hline\hline

\multirow{3}{*}{-Inc-v3\textsubscript{ens3}} & FGSM & 36.1 & 15.4 \\
& I-FGSM & \textbf{99.6} & 18.6 \\
& MI-FGSM & \textbf{99.6} & \textbf{37.6} \\
\hline
\multirow{3}{*}{-Inc-v3\textsubscript{ens4}} & FGSM & 33.0 & 15.0 \\
& I-FGSM & 99.2 & 18.7 \\
& MI-FGSM & \textbf{99.3} & \textbf{40.3} \\
\hline
\multirow{3}{*}{-IncRes-v2\textsubscript{ens}} & FGSM & 36.2 & 6.4 \\
& I-FGSM & 99.5 & 9.9 \\
& MI-FGSM & \textbf{99.7} & \textbf{23.3} \\

\hline
\end{tabular}
\end{center}
\vspace{-2ex}
\caption{The success rates (\%) of non-targeted adversarial attacks against an ensemble of white-box models and a hold-out black-box target model. We include seven models---Inc-v3, Inc-v4, IncRes-v2, Res-152, Inc-v3\textsubscript{ens3}, Inc-v3\textsubscript{ens4} and IncRes-v2\textsubscript{ens}. In each row, ``-'' indicates the name of the hold-out model and the adversarial examples are generated for the ensemble of the other six models.}
\label{tab:ensemble-models}
\vspace{-2ex}
\end{table}
To attack the adversarially trained models in the black-box manner, we include all seven models introduced in Sec.~\ref{sec:setup}. Similarly, we keep one adversarially trained model as the hold-out target model to evaluate the performance in the black-box manner, and attack the rest six model in an ensemble, whose logits are fused together with equal ensemble weights. The perturbation $\epsilon$ is $16$ and the decay factor $\mu$ is $1.0$. We compare the results of FGSM, I-FGSM and MI-FGSM with $20$ iterations. The results are shown in Table~\ref{tab:ensemble-models}.

It can be seen that the adversarially trained models also cannot defend our attacks effectively, which can fool Inc-v3\textsubscript{ens4} by more than $40\%$ of the adversarial examples. 
Therefore, the models obtained by ensemble adversarial training, the most robust models trained on the ImageNet as far as we are concerned, are vulnerable to our attacks in the black-box manner, thus causing new security issues for developing algorithms to learn robust deep learning models.

\subsection{Competitions}
\label{sec:competition}

There are three sub-competitions in the NIPS 2017 Adversarial Attacks and Defenses Competition, which are the Non-targeted Adversarial Attack, Targeted Adversarial Attack and Defense Against Adversarial Attack. The organizers provide $5000$ ImageNet-compatible images for evaluating the attack and defense submissions. For each attack, one adversarial example is generated for each image with the size of perturbation ranging from $4$ to $16$ (specified by the organizers), and all adversarial examples run through all defense submissions to get the final score. We won the first places in both the non-targeted attack and targeted attack by the method introduced in this paper. We will specify the configurations in our submissions.

For the non-targeted attack\footnote{Source code is available at \url{https://github.com/dongyp13/Non-Targeted-Adversarial-Attacks}.}, we implement the MI-FGSM for attacking an ensemble of Inc-v3, Inc-v4, IncRes-v2, Res-152, Inc-v3\textsubscript{ens3}, Inc-v3\textsubscript{ens4}, IncRes-v2\textsubscript{ens} and Inc-v3\textsubscript{adv}~\cite{kurakin2017adversarial}. We adopt the ensemble in logits scheme. The ensemble weights are set as $\nicefrac{1}{7.25}$ equally for the first seven models and $\nicefrac{0.25}{7.25}$ for Inc-v3\textsubscript{adv}. The number of iterations is $10$ and the decay factor $\mu$ is $1.0$.

For the targeted attack\footnote{Source code is available at \url{https://github.com/dongyp13/Targeted-Adversarial-Attacks}.}, we build two graphs for attacks. If the size of perturbation is smaller than $8$, we attack Inc-v3 and IncRes-v2\textsubscript{ens} with ensemble weights $\nicefrac{1}{3}$ and $\nicefrac{2}{3}$; otherwise we attack an ensemble of Inc-v3, Inc-v3\textsubscript{ens3}, Inc-v3\textsubscript{ens4}, IncRes-v2\textsubscript{ens} and Inc-v3\textsubscript{adv} with ensemble weights $\nicefrac{4}{11}, \nicefrac{1}{11}, \nicefrac{1}{11}, \nicefrac{4}{11}$ and $\nicefrac{1}{11}$. The number of iterations is $40$ and $20$ respectively, and the decay factor $\mu$ is also $1.0$.

\section{Discussion}
Taking a different perspective, we think that finding an adversarial example is an analogue to training a model and the transferability of the adversarial example is also an analogue to the generalizability of the model. By taking a meta view, we actually ``train'' an adversarial example given a set of models as training data. In this way, the improved transferability obtained by the momentum and ensemble methods is reasonable because the generalizability of a model is usually improved by adopting the momentum optimizer or training on more data. And we think that other tricks (\eg, SGD) for enhancing the generalizability of a model could also be incorporated into adversarial attacks for better transferability.

\section{Conclusion}

In this paper, we propose a broad class of momentum-based iterative methods to boost adversarial attacks, which can effectively fool white-box models as well as black-box models.
Our methods consistently outperform one-step gradient-based methods and vanilla iterative methods in the black-box manner.
We conduct extensive experiments to validate the effectiveness of the proposed methods and explain why they work in practice.
To further improve the transferability of the generated adversarial examples, we propose to attack an ensemble of models whose logits are fused together. 
We show that the models obtained by ensemble adversarial training are vulnerable to our black-box attacks, which raises new security issues for the development of more robust deep learning models.
\vspace{-1ex}
\section*{Acknowledgements}
\vspace{-1ex}
\small{
The work is supported by the National NSF of China (Nos. 61620106010, 61621136008, 61332007, 61571261 and U1611461), Beijing Natural Science Foundation (No. L172037), Tsinghua Tiangong Institute for Intelligent Computing and the NVIDIA NVAIL Program, and partially funded by Microsoft Research Asia and Tsinghua-Intel Joint Research Institute.
}

{\small
\bibliographystyle{ieee}
\bibliography{egbib}
}

\newpage
\clearpage
\label{sec:appendix}
\noindent \begin{center} {\large  \textbf{Appendix}} \end{center}
\setcounter{section}{0}
\renewcommand\thesection{\Alph{section}}

In this supplementary material, we provide more results in our experiments. In Sec.~\ref{sec:l2}, we report the success rates of non-targeted attacks based on $L_2$ norm bound. In Sec.~\ref{sec:target}, we provide the results of targeted attacks. The experiements consistently demonstrate the effectiveness of the proposed momentum-based methods.

\section{Non-targeted attacks based on $\bm{L_{2}}$ norm bound}
\label{sec:l2}
We first perform non-targeted attacks based on $L_2$ norm bound. Since the $L_2$ distance between an adversarial example and a real example is defined as
\begin{equation}
\|\bm{x}^* - \bm{x}\|_2 = \sqrt{\sum_{i=1}^{N}(x^*_i - x_i)^2},
\end{equation}
where $N$ is the dimension of input $\bm{x}$ and $\bm{x}^*$, the distance measure depends on $N$. For example, if the distance of each dimension of an adversarial example and a real example is $|x^*_i - x_i| = \epsilon$, the $L_2$ norm is $\epsilon\sqrt{N}$ between them while the $L_{\infty}$ norm is $\epsilon$. Therefore, we set the $L_2$ norm bound as $16\sqrt{N}$ in our experiments, where $N$ is the dimension of the input to a network.

\begin{table}[!thb]
\begin{center}
\begin{tabular}{c|c|c|c}

\hline
& Attack & Ensemble & Hold-out \\
\hline\hline

\multirow{3}{*}{-Inc-v3} & FGM & 47.3 & 52.7\\
& I-FGM & 99.1 & 65.3\\
& MI-FGM & \textbf{99.2} & \textbf{89.7}\\
\hline
\multirow{3}{*}{-Inc-v4} & FGM & 47.2 & 49.3 \\
& I-FGM & 99.3 & 56.7 \\
& MI-FGM & \textbf{99.4} & \textbf{88.0} \\
\hline
\multirow{3}{*}{-IncRes-v2} & FGSM & 47.3 & 50.4 \\
& I-FGSM & 99.4 & 54.3 \\
& MI-FGSM & \textbf{99.5} & \textbf{86.1} \\
\hline
\multirow{3}{*}{-Res-152} & FGM & 47.6 & 46.6 \\
& I-FGM & 99.0 & 44.7 \\
& MI-FGM & \textbf{99.5} & \textbf{81.4} \\
\hline\hline
\multirow{3}{*}{-Inc-v3\textsubscript{ens3}} & FGM & 51.8 & 35.4 \\
& I-FGM & \textbf{99.8} & 29.5 \\
& MI-FGM & 99.6 & \textbf{59.8} \\
\hline
\multirow{3}{*}{-Inc-v3\textsubscript{ens4}} & FGM & 51.2 & 37.5 \\
& I-FGM & 99.2 & 36.4 \\
& MI-FGM & \textbf{99.7} & \textbf{66.5} \\
\hline
\multirow{3}{*}{-IncRes-v2\textsubscript{ens}} & FGSM & 54.4 & 32.4 \\
& I-FGSM & 99.2 & 19.9 \\
& MI-FGSM & \textbf{99.8} & \textbf{56.4} \\

\hline
\end{tabular}
\end{center}
\vspace{-2ex}
\caption{The success rates (\%) of non-targeted adversarial attacks based on $L_2$ norm bound against an ensemble of white-box models and a hold-out black-box target model. In each row, ``-'' indicates the name of the hold-out model and the adversarial examples are generated for the ensemble of the other six models.}
\label{tab:ensemble-models-l2}
\end{table}

\subsection{Attacking a single model}

We include seven networks in this section, which are Inc-v3, Inc-v4, IncRes-v2, Res-152, Inc-v3\textsubscript{ens3}, Inc-v3\textsubscript{ens4} and IncRes-v2\textsubscript{ens}. We generate adversarial examples for Inc-v3, Inc-v4, IncRes-v2 and Res-152 respectively, and measure the success rates of attacks on all models. We compare three attack methods, which are the fast gradient method (FGM, defined in Eq.~\eqref{eq:fg}), iterative FGM (I-FGM) and momentum iterative FGM (MI-FGM, defined in Eq.~\eqref{eq:mi-fgs}). We set the number of iterations to $10$ in I-FGM and MI-FGM, and the decay factor to $1.0$ in MI-FGM.

The results are shown in Table~\ref{tab:single-model-l2} (See next page.). We can also see that MI-FGM attacks a white-box model with a near $100\%$ success rate as I-FGM, and outperforms FGM and I-FGM in black-box attacks significantly. The conclusions are similar to those of $L_{\infty}$ norm bound experiments, which consistently demonstrate the effectiveness of the proposed momentum-based iterative methods.

\subsection{Attacking an ensemble of models}
\label{sec:l2-ens}

\begin{table*}[t]
\begin{center}
\begin{tabular}{c|c|c|c|c|c|c|c|c}

\hline
& Attack & Inc-v3 & Inc-v4 & IncRes-v2 & Res-152 & Inc-v3\textsubscript{ens3} & Inc-v3\textsubscript{ens4} & IncRes-v2\textsubscript{ens} \\
\hline\hline

\multirow{3}{*}{Inc-v3} & FGM & 76.2$^*$ & 41.0 & 43.1 & 41.3 & 34.6 & 34.9 & 26.2\\
& I-FGM & \textbf{100.0}$^*$ & 39.9 & 36.4 & 27.5 & 17.5 & 19.2 & 10.9\\
& MI-FGM & \textbf{100.0}$^*$ & \textbf{67.6} & \textbf{66.3} & \textbf{56.1} & \textbf{44.4} & \textbf{45.5} & \textbf{33.9}\\
\hline
\multirow{3}{*}{Inc-v4} & FGM & 47.3 & 63.1$^*$ & 37.3 & 39.0 & 35.3 & 33.9 & 27.7 \\
& I-FGM & 52.8 & \textbf{100.0}$^*$ & 42.0 & 33.5 & 21.9 & 19.9 & 13.8 \\
& MI-FGM & \textbf{76.9} & \textbf{100.0}$^*$ & \textbf{69.6} & \textbf{59.7} & \textbf{51.2} & \textbf{51.0} & \textbf{39.4} \\
\hline
\multirow{3}{*}{IncRes-v2} & FGM & 48.2 & 38.9 & 60.4$^*$ & 39.8 & 36.6 & 35.5 & 30.5 \\
& I-FGM & 56.0 & 47.5 & \textbf{99.6}$^*$ & 36.9 & 27.5 & 22.9 & 18.7 \\
& MI-FGM & \textbf{81.7} & \textbf{75.8} & \textbf{99.6}$^*$ & \textbf{66.9} & \textbf{62.7} & \textbf{57.7} & \textbf{58.8} \\
\hline
\multirow{3}{*}{-Res-152} & FGM & 50.8 & 40.7 & 42.0 & 75.1$^*$ & 36.5 & 36.0 & 31.6\\
& I-FGM & 47.6 & 43.9 & 43.9 & 99.4$^*$ & 32.7 & 32.3 & 25.2 \\
& MI-FGM & \textbf{71.3} & \textbf{65.5} & \textbf{64.3} & \textbf{99.6}$^*$ & \textbf{56.7} & \textbf{55.4} & \textbf{51.5} \\

\hline
\end{tabular}
\end{center}
\vspace{-2ex}
\caption{The success rates (\%) of non-targeted adversarial attacks based on $L_2$ norm bound against all models. The adversarial examples are crafted for Inc-v3, Inc-v4, IncRes-v2 and Res-152 respectively using FGM, I-FGM and MI-FGM. $^*$ indicates the white-box attacks.}
\label{tab:single-model-l2}
\end{table*}
In this experiments, we also include Inc-v3, Inc-v4, IncRes-v2, Res-152, Inc-v3\textsubscript{ens3}, Inc-v3\textsubscript{ens4} and IncRes-v2\textsubscript{ens} models for our study. We keep one model as the hold-out black-box model and attack an ensemble of the other six models by FGM, I-FGM and MI-FGM respectively. We set the number of iterations to $20$ in I-FGM and MI-FGM, the decay factor to $1.0$ in MI-FGM, and the ensemble weights to $\nicefrac{1}{6}$ equally.

We show the results in Table~\ref{tab:ensemble-models-l2}. Iterative methods including I-FGM and MI-FGM can obtain a near $100\%$ success rate for an ensemble of white-box models. And MI-FGM can attack a black-box model with a much higher success rate, showing the good transferability of the adversarial examples generated by MI-FGM. For adversarially trained models, MI-FGM can fool them with about $60\%$ success rates, revealing the great vulnerability of the adversarially trained models against our black-box attacks.

\section{Targeted attacks}
\label{sec:target}

\subsection{$\bm{L_{\infty}}$ norm bound}

\begin{table}
\begin{center}
\begin{tabular}{c|c|c|c}

\hline
& Attack & Ensemble & Hold-out \\
\hline\hline

\multirow{3}{*}{-Inc-v3} & FGSM & 0.5 & 0.5\\
& I-FGSM & \textbf{99.6} & 9.0\\
& MI-FGSM & 99.5 & \textbf{17.6}\\
\hline
\multirow{3}{*}{-Inc-v4} & FGSM & 0.3 & 0.4 \\
& I-FGSM & \textbf{99.9} & 7.0 \\
& MI-FGSM & 99.8 & \textbf{15.6} \\
\hline
\multirow{3}{*}{-IncRes-v2} & FGSM & 0.4 & 0.2 \\
& I-FGSM & \textbf{99.9} & 7.3 \\
& MI-FGSM & 99.8 & \textbf{16.1} \\
\hline
\multirow{3}{*}{-Res-152} & FGSM & 0.1 & 0.5 \\
& I-FGSM & \textbf{99.6} & 3.3 \\
& MI-FGSM & 99.5 & \textbf{11.4} \\
\hline\hline
\multirow{3}{*}{-Inc-v3\textsubscript{ens3}} & FGSM & 0.3 & 0.1 \\
& I-FGSM & \textbf{99.7} & 0.1 \\
& MI-FGSM & \textbf{99.7} & \textbf{0.5} \\
\hline
\multirow{3}{*}{-Inc-v3\textsubscript{ens4}} & FGSM & 0.2 & 0.1 \\
& I-FGSM & \textbf{99.9} & 0.4 \\
& MI-FGSM & 99.8 & \textbf{0.9} \\
\hline
\multirow{3}{*}{-IncRes-v2\textsubscript{ens}} & FGSM & 0.5 & 0.1 \\
& I-FGSM & 99.7 & 0.1 \\
& MI-FGSM & \textbf{99.8} & \textbf{0.2} \\

\hline
\end{tabular}
\end{center}
\caption{The success rates (\%) of targeted adversarial attacks based on $L_{\infty}$ norm bound against an ensemble of white-box models and a hold-out black-box target model. In each row, ``-'' indicates the name of the hold-out model and the adversarial examples are generated for the ensemble of the other six models.}
\label{tab:target}
\end{table}
\begin{table} [!t]
\begin{center}
\begin{tabular}{c|c|c|c}

\hline
& Attack & Ensemble & Hold-out \\
\hline\hline

\multirow{3}{*}{-Inc-v3} & FGM & 0.7 & 0.4\\
& I-FGM & \textbf{99.7} & 17.8\\
& MI-FGM & 99.5 & \textbf{21.0}\\
\hline
\multirow{3}{*}{-Inc-v4} & FGM & 0.7 & 0.5 \\
& I-FGM & \textbf{99.9} & 15.2 \\
& MI-FGM & 99.8 & \textbf{21.8} \\
\hline
\multirow{3}{*}{-IncRes-v2} & FGM & 0.7 & 0.7 \\
& I-FGM & 99.8 & 16.4 \\
& MI-FGM & \textbf{99.9} & \textbf{21.7} \\
\hline
\multirow{3}{*}{-Res-152} & FGM & 0.5 & 0.4 \\
& I-FGM & 99.5 & 9.2 \\
& MI-FGM & \textbf{99.6} & \textbf{17.4} \\
\hline\hline
\multirow{3}{*}{-Inc-v3\textsubscript{ens3}} & FGM & 0.6 & 0.2 \\
& I-FGM & \textbf{99.9} & 0.7 \\
& MI-FGM & 99.6 & \textbf{1.6} \\
\hline
\multirow{3}{*}{-Inc-v3\textsubscript{ens4}} & FGM & 0.5 & 0.2 \\
& I-FGM & 99.7 & 1.7 \\
& MI-FGM & \textbf{100.0} & \textbf{2.0} \\
\hline
\multirow{3}{*}{-IncRes-v2\textsubscript{ens}} & FGM & 0.6 & 0.4 \\
& I-FGM & 99.6 & 0.5 \\
& MI-FGM & \textbf{99.8} & \textbf{1.9} \\

\hline
\end{tabular}
\end{center}
\caption{The success rates (\%) of targeted adversarial attacks based on $L_2$ norm bound against an ensemble of white-box models and a hold-out black-box target model. In each row, ``-'' indicates the name of the hold-out model and the adversarial examples are generated for the ensemble of the other six models.}
\label{tab:target-l2}
\end{table}

Targeted attacks are much more difficult than non-targeted attacks in the black-box manner, since they require the black-box model to output the specific target label. For DNNs trained on a dataset with thousands of output categories such as the ImageNet dataset, finding targeted adversarial examples by only one model to fool a black-box model is impossible~\cite{liu2016delving}. Thus we perform targeted attacks by integrating the ensemble-based approach.

We show the results in Table~\ref{tab:target}, where the success rate is measured by the percentage of the adversarial examples that are classified as the target label by the model. 
Similar to the experimental settings in Sec.~\ref{sec:ensemble-non-targeted}, we keep one model to test the performance of black-box attacks, with the targeted adversarial examples generated for the ensemble of the other six models. We set the size of perturbation $\epsilon$ to $48$, decay factor $\mu$ to $1.0$ and the number of iterations to $20$ for I-FGSM and MI-FGSM. We can see that one-step FGSM can hardly attack the ensemble of models as well as the target black-box models. The success rates of the adversarial examples generated by MI-FGSM are close to $100\%$ for white-box models and higher than $10\%$ for normally trained black-box models. Unfortunately, it cannot effectively generate targeted adversarial examples to fool adversarially trained models, which remains an open issue for future researches.

\subsection{$\bm{L_2}$ norm bound}

We draw similar conclusions for targeted attacks based on $L_2$ norm bound. In our experiments, we also include Inc-v3, Inc-v4, IncRes-v2, Res-152, Inc-v3\textsubscript{ens3}, Inc-v3\textsubscript{ens4} and IncRes-v2\textsubscript{ens} models. We keep one model as the hold-out black-box model and attack an ensemble of the other six models with equal ensemble weights by FGM, I-FGM and MI-FGM respectively. We set the maximum perturbation $\epsilon$ to $48\sqrt{N}$ where $N$ is the dimension of inputs, the number of iterations to $20$ in I-FGM and MI-FGM, and the decay factor to $1.0$ in MI-FGM. We report the success rates of adversarial examples against the white-box ensemble of models and the black-box target model in Table~\ref{tab:target-l2}. MI-FGM can easily fool white-box models, but it cannot fool the adversarially trained models effectively in the targeted black-box attacks.

\end{document}